\documentclass[letterpaper, 10 pt, conference]{ieeeconf}
\IEEEoverridecommandlockouts
\usepackage{amsmath,amsfonts,amssymb}
\usepackage{algorithmic}
\usepackage{algorithm}
\usepackage{array}
\usepackage[caption=false,font=normalsize,labelfont=sf,textfont=sf]{subfig}
\usepackage{textcomp}
\usepackage{stfloats}
\usepackage{url}
\usepackage{verbatim}
\usepackage{graphicx}
\usepackage{cite}
\usepackage{capt-of}  % <---
\usepackage{cuted}    % <===
\usepackage{multirow}
\usepackage{placeins, afterpage}
\usepackage{float}
\usepackage{cuted}
\usepackage{xcolor}
\usepackage{multicol}
\usepackage{mathtools}
\usepackage{lipsum}
\usepackage{float}
\usepackage{wrapfig}
\usepackage{orcidlink}
\usepackage[inkscapeformat=png]{svg}

\begin{document}
\title{Sliding Sensors: Configurable Confidence in State Estimation \\ for Continuum Robots}

\author{Ella Walsh$^*$, Spencer Teetaert$^*$, Eric Diller, Timothy D. Barfoot, Jessica Burgner-Kahrs
\thanks{$^*$Equal contribution.}
\thanks{The authors are with the University of Toronto Robotics Institute, Toronto, ON, Canada. E-mail: ella.walsh@mail.utoronto.ca.}
\thanks{This work was supported in part by the National Sciences and Engineering Research Council of Canada (NSERC) and the Queen Elizabeth II Graduate Scholarship in Science and Technology (QEII‐GSST).}% <-this % stops a space
}

\maketitle

\section{Introduction}\label{sec:introduction}

Continuum robots often operate in uncertain environments, where accurate state estimation is essential for safe interactions. Estimate uncertainty is inherently spatially non-uniform: confidence varies depending on where measurements are available. Global estimation accuracy is not always the top priority, but rather achieving sufficient confidence at task-relevant locations along the robot.

Sensor placement and state estimation for continuum robots are fundamentally coupled~\cite{mahoney2016}. Existing works based on the paradigm of optimal sensor placement investigate formulations that maximize estimation speed and/or accuracy under constraints on the quantity or locations of sensors~\cite{mahoney2016, kim2014}. Such constraints lead to regions of low confidence in state estimates, where measurements do not directly provide information about the state of a robot at a given point~\cite{Anderson2017, lilge2022, ferguson2024}. Rather, robot models are required to infer what the state may be in between sensor locations, leading to increased uncertainty further away from measurement points. 

A limitation of a fixed-placement sensor design arises when task-relevant measurement points change during operation. Many tasks require moving zones of interest, such as a robot moving through a cluttered environment while detecting or avoiding contact~\cite{li2017}, or coordinating the motion or connection of multiple collaborative continuum robots~\cite{Li2024, Anderson2017, Russo2022}. In such cases, the optimal sensor location is not fixed along the robot, and it would be advantageous to move measurement points toward external regions of interest. Some works show this capability passively, with tip-fixed sensors that have some motion relative to the robot backbone as the robot bends or extends~\cite{lin2022, Lu2022, Donder2022}. However, to the best of our knowledge, no work explores sensor motion as an actuated degree of freedom or its implications in state estimation.

This work introduces mechanically reconfigurable sensing enabling uncertainty-shaping in state estimation for continuum robots. We present a concept hardware design demonstrating the feasibility of longitudinal translation of a sensor within a continuum robot. 

We demonstrate that state estimation confidence can be reconfigured by varying the sensor location, and show a reduction of full-body shape estimation errors when sliding the sensor back and forth over time, compared to a single fixed tip sensor.

\begin{figure}[!t]
    \includegraphics[width=\columnwidth]{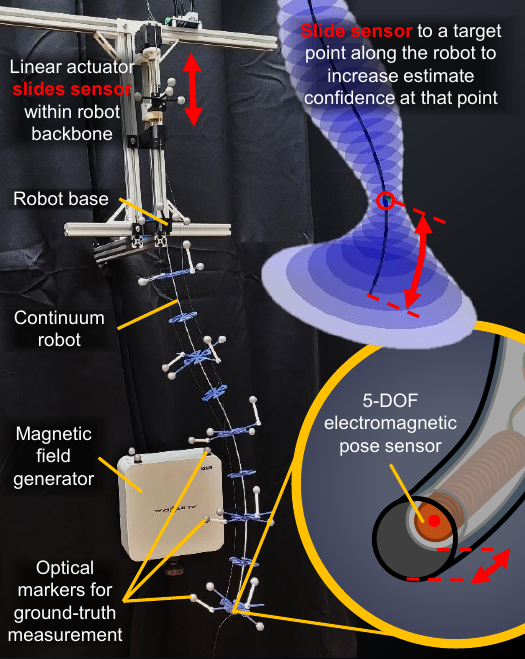}
    \caption{Proposed method of sliding a sensor within a continuum robot to control the location of high-confidence estimate regions.}
    \label{fig:title_fig}
\end{figure}

\section{Methodology}\label{sec:methodology}

\subsection{Hardware}

The system consists of a sensor mounted in an inner tube, linearly actuated within an outer tube, making up the backbone of a continuum robot. Any sensor can be used so long as it can fit within the backbone of the robot of interest, and the linear actuation method can be substituted for any method achieving the desired stroke length. 

For proof-of-concept in this work, we use a 0.5 mm diameter, 8 mm length electromagnetic coil 5-DOF position sensor with an external magnetic field generator (Aurora v3, Northern Digital Inc., Canada). The linear actuator consists of a lead screw system with 60 mm stroke length (Igus Inc., Germany). The linear actuation is driven by a Dynamixel XL430-W250 servo motor for precise position control, which is mapped to the translation of the sensor tube to measure the length of sensor displacement. The inner tube (0.8 mm ID, 1.2 mm OD) and the outer tube (1.8 mm ID, 2.0 mm OD) are both made of nitinol, and the tendon-driven robot is 70 cm long.

During operation for a single trial under a static robot pose, the sensor is oscillated three full cycles with maximum stroke length at a speed of 24 mm/s. Ground-truth frame measurements for the robot and the sliding tube are collected via optical trackers (Vicon Motion Systems Ltd., UK). The mechanical setup can be seen in Figure \ref{fig:title_fig}.

\subsection{Estimation}
Estimation is formulated as a factor graph optimization problem. Sliding the sensor adds time-varying measurement factors at varying arclength coordinates. As a prior model, we choose to use the estimation method of~\cite{Teetaert2025a, Teetaert2025b} as it enables continuous representations in both arclength and time. The state that we estimate includes the pose, strain, and velocity of the robot. We model the 5-DOF pose sensor measurement through the $SE(3)$ distance metric:
\begin{align}
    \nonumber \boldsymbol{e}_m(s, t) &= {\ln(\tilde{\boldsymbol{T}}^{-1}_m \boldsymbol{T}(s, t))}^\vee.
\end{align} For a 5-DOF measurement error, we project out the yaw component by removing the last term in the error vector $\boldsymbol{e}_m$, making use of a small angle approximation. This error is used along with a Gaussian noise model as a measurement factor in the estimator factor graph.

\section{RESULTS}\label{sec:results}

\begin{figure}[t!]\label{fig:side_by_side}
    \centering
    \includegraphics[width=\columnwidth]{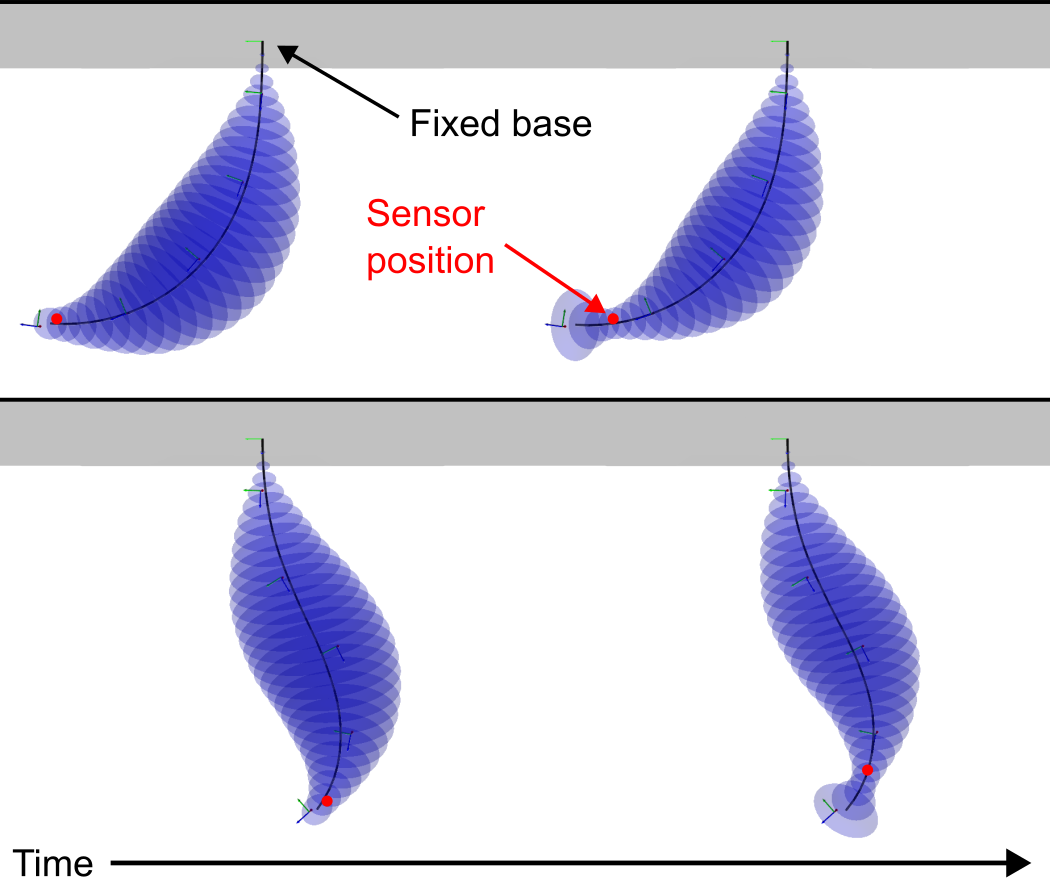}
    \caption{Robot position estimate mean (black line) with $3\sigma$ confidence ellipsoids (blue) is shown. Three configurations are shown: a straight (left), C-shape (middle), and S-shape (right) robot. For each shape, results are shown with a 5-DOF pose sensor at the tip and translated 6 cm into the robot. Ground-truth robot poses from the motion capture system are drawn as coordinate frames at approximately 12 cm intervals along the robot. }
\end{figure}
\begin{figure}[t!]\label{fig:sim_results}
    \centering
    \includegraphics[width=\columnwidth]{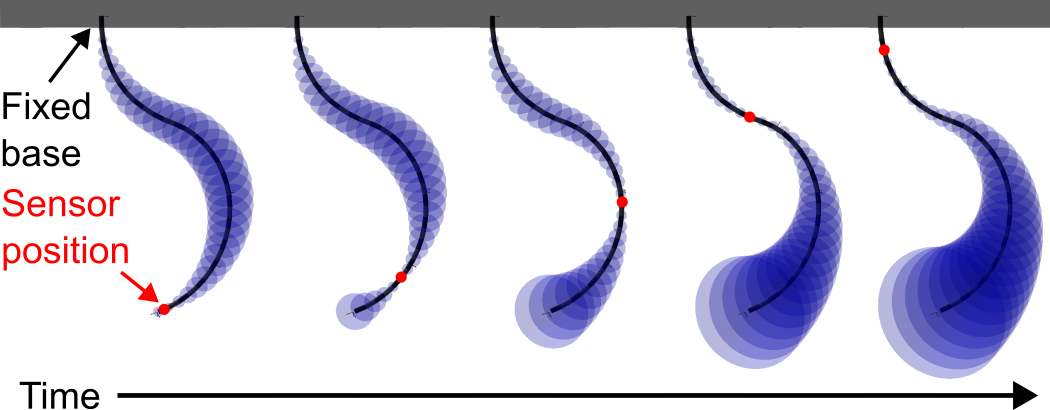}
    \caption{Robot position estimate mean (black line) with $3\sigma$ confidence ellipsoids (blue) is shown. Data is generated from a simulated continuum robot, with a 5-DOF pose sensor mounted on a sliding mechanism that can move down the entire robot length. Several timesteps are shown as the sensor moves down the robot.}
\end{figure}

\begin{table}[ht!]\label{tab:error-table}
\centering
\caption{Shape estimation errors comparing the fixed and oscillating sensor configuration on real-world data.}
\begin{tabular}{lll}
\hline
 & C-shape robot & S-shape robot \\
\hline
\multicolumn{3}{l}{\textit{Position RMSE (cm)}} \\
Fixed sensor & 1.64 & 1.78 \\
Sliding sensor & 1.54 \textit{(-6.1\%)} & 1.63 \textit{(-8.4\%)} \\
\hline
\multicolumn{3}{l}{\textit{Orientation RMSE (rad)}} \\
Fixed sensor  & 0.0779 & 0.2381 \\
Sliding sensor  &  0.0754 \textit{(-3.2\%)} & 0.2291 \textit{(-3.8\%)} \\
\hline
\end{tabular}
\end{table}

The robot shape estimates and corresponding uncertainty profiles for three different robot shapes are shown in Figure~\ref{fig:side_by_side}. Similarly, estimates for a simulated robot with a sensor sliding the full robot length can be seen in Figure~\ref{fig:sim_results}. It can be seen that the location of relatively higher confidence near the tip of the robot matches the location of the 5-DOF sensor within the robot. The results in Table~\ref{tab:error-table} show that sliding the sensor back and forth over time leads to minor improvement in shape estimation accuracy---up to 8.4\% position error reduction and 3.8\% orientation error reduction---compared to the case where the sensor is fixed at the tip. We hypothesize that the motion prior used during estimation, alongside the moving measurement, are able to leverage sensor information temporally to produce more accurate results. This is expected to be more significant when the sensor is moved fast relative to the motion of the robot, though further study is required to confirm these claims. 

\section{DISCUSSION}\label{sec:discussion}
Our work demonstrates the feasibility of sliding a sensor along a continuum robot to configure confidence bounds in the state estimate. 

We plan to continue exploring design solutions for active sensing including overcoming the strict sensor size requirements to be contained within a central backbone. Furthermore, this work only analyzes static and quasi-static robot configurations; future work will explore a dynamic robot and the effects of sensor dynamics such as sensor sliding frequency.

This approach unlocks a new paradigm of active estimation in continuum robotics, laying foundations for future work in belief-aware planning, where both actuation and sensing configurations are jointly optimized. We hope this work will inspire future research into the use of sliding sensors in continuum robots.

\newpage
\bibliographystyle{IEEEtran}
\bibliography{references}

\end{document}